\documentclass[10pt,twocolumn,letterpaper]{article}

\usepackage{iccv}
\usepackage{times}
\usepackage{epsfig}
\usepackage{graphicx}
\usepackage{amsmath}
\usepackage{amssymb}

\usepackage[breaklinks=true,bookmarks=false]{hyperref}

\iccvfinalcopy 

\def\iccvPaperID{****} 

\newcommand{\rev}[1]{{#1}}

\ificcvfinal\fi
\begin{document}

\title{Similarity-Preserving Knowledge Distillation}

\author{Frederick Tung$^{1,2}$ and Greg Mori$^{1,2}$\\
$^1$Simon Fraser University \quad $^2$Borealis AI\\
{\tt\small ftung@sfu.ca, mori@cs.sfu.ca}
}

\maketitle

\begin{abstract}
Knowledge distillation is a widely applicable technique for training a student neural network under the guidance of a trained teacher network. For example, in neural network compression, a high-capacity teacher is distilled to train a compact student; in privileged learning, a teacher trained with privileged data is distilled to train a student without access to that data. The distillation loss determines how a teacher's knowledge is captured and transferred to the student. 
In this paper, we propose a new form of knowledge distillation loss that is inspired by the observation that semantically similar inputs tend to elicit similar activation patterns in a trained network. Similarity-preserving knowledge distillation guides the training of a student network such that input pairs that produce similar (dissimilar) activations in the teacher network produce similar (dissimilar) activations in the student network. In contrast to previous distillation methods, the student is not required to mimic the representation space of the teacher, but rather to preserve the pairwise similarities in its own representation space. Experiments on three public datasets demonstrate the potential of our approach.
\end{abstract}

\section{Introduction}

\begin{figure}[t]
\centering
\includegraphics[width=\linewidth, trim=20 140 220 20, clip=true]{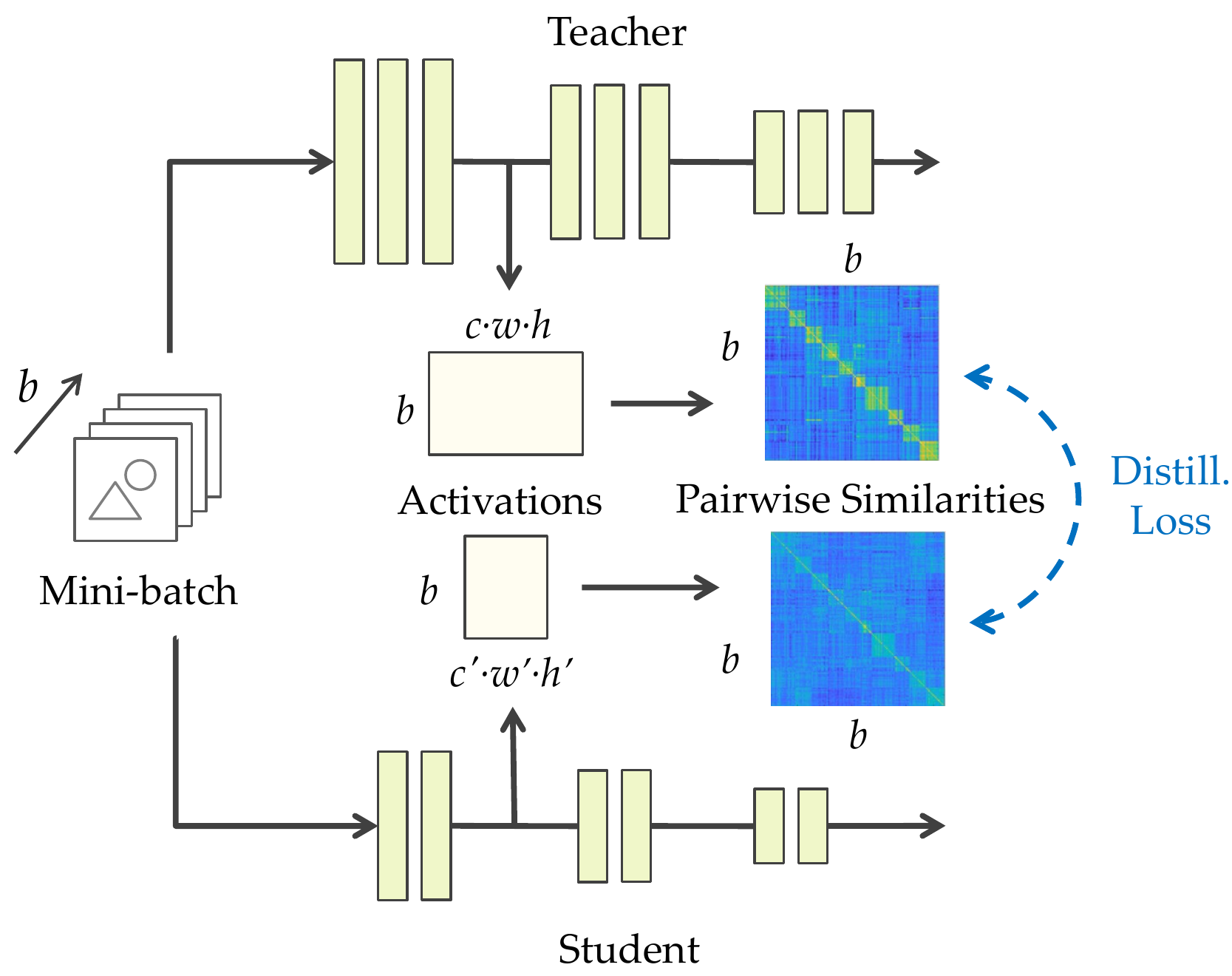}
\caption{\textit{Similarity-preserving knowledge distillation} guides the training of a student network such that input pairs that produce similar (dissimilar) activations in the pre-trained teacher network produce similar (dissimilar) activations in the student network. Given an input mini-batch of $b$ images, we derive $b \times b$ pairwise similarity matrices from the activation maps, and compute a distillation loss on the matrices produced by the student and the teacher.} 
\label{fig:pull}
\end{figure}

Deep neural networks are being used to solve an increasingly wide array of computer vision problems. While the general trend in deep learning is towards deeper, wider, and more complex networks, deploying deep learning solutions in the real world requires us to consider the computational cost. A mobile robot or self-driving vehicle, for example, has limited memory and power. Even when resources are abundant, such as when a vision system is hosted in the cloud, more resource-efficient deep networks mean more clients can be served at a lower cost. 
When performing transfer learning in the real world, data privilege and privacy issues may restrict access to data in the source domain. It may be necessary to transfer the knowledge of a network trained on the source domain assuming access only to training data from the target task domain.

Knowledge distillation is a general technique for supervising the training of ``student" neural networks by capturing and transferring the knowledge of trained ``teacher" networks. While originally motivated by the task of neural network compression for resource-efficient deep learning \cite{hintonetal2015}, knowledge distillation has found wider applications in such areas as privileged learning \cite{lopezpazetal2016}, adversarial defense \cite{papernotetal2016}, and learning with noisy data \cite{lietal2017}. Knowledge distillation is conceptually simple: it guides the training of a student network with an additional distillation loss that encourages the student to mimic some aspect of a teacher network. Intuitively, the trained teacher network provides a richer supervisory signal than the data supervision (e.g. annotated class labels) alone.

The conceptual simplicity of knowledge distillation belies the fact that \textit{how} to best capture the knowledge of the teacher to train the student (i.e. how to define the distillation loss) remains an open question. In traditional knowledge distillation \cite{hintonetal2015}, the softened class scores of the teacher are used as the extra supervisory signal: the distillation loss encourages the student to mimic the scores of the teacher. FitNets \cite{romeroetal2015} extend this idea by adding hints to guide the training of intermediate layers.
In flow-based knowledge distillation \cite{yimetal2017}, the extra supervisory signal comes from the inter-layer ``flow" -- how features are transformed between layers. The distillation loss encourages the student to mimic the teacher's flow matrices, which are derived from the inner product between feature maps in two layers, such as the first and last layers in a residual block. 
In attention transfer \cite{zagoruykokomodakis2017}, the supervisory signal for knowledge distillation is in the form of spatial attention. Spatial attention maps are computed by summing the squared activations along the channel dimension. The distillation loss encourages the student to produce similar normalized spatial attention maps as the teacher, intuitively paying attention to similar parts of the image as the teacher.

In this paper, we present a novel form of knowledge distillation that is inspired by the observation that semantically similar inputs tend to elicit similar activation patterns in a trained neural network. Similarity-preserving knowledge distillation guides the training of a student network such that input pairs that produce similar (dissimilar) activations in the trained teacher network produce similar (dissimilar) activations in the student network. Figure \ref{fig:pull} shows the overall procedure. Given an input mini-batch of $b$ images, we compute pairwise similarity matrices from the output activation maps. The $b \times b$ matrices encode the similarities in the activations of the network as elicited by the images in the mini-batch. Our distillation loss is defined on the pairwise similarity matrices produced by the student and the teacher.

\begin{figure}
\centering
\includegraphics[width=0.9\linewidth, trim=100 180 100 220, clip=true]{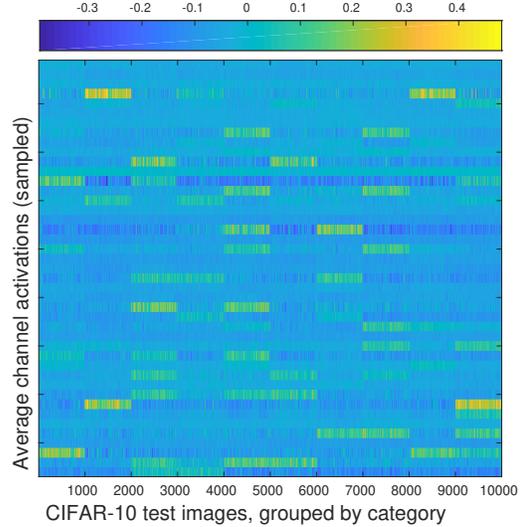}
\caption{Semantically similar inputs tend to elicit similar activation patterns in a trained neural network. This visualization shows channel-wise average activations sampled from the last convolutional layer of a WideResNet-16-2 network on the CIFAR-10 test images. Activation patterns are largely consistent within the same category (e.g. columns 1 to 1000) and distinctive across different categories (e.g. columns 1 to 1000 vs. columns 1001 to 2000).}
\label{fig:activations}
\end{figure}

To support the intuition of our distillation loss, Figure \ref{fig:activations} visualizes the average activation of each channel in the last convolutional layer of a WideResNet-16-2 teacher network (we adopt the standard notation WideResNet-$d$-$k$ to refer to a wide residual network \cite{zagoruykokomodakis2016} with depth $d$ and width multiplier $k$), on the CIFAR-10 test images. We can see that images from the same object category tend to activate similar channels in the trained network. The similarities in activations across different images capture useful semantics learned by the teacher network. We study whether these similarities provide an informative supervisory signal for knowledge distillation. 

The contributions of this paper are:
\begin{itemize}
    \item We introduce \textit{similarity-preserving knowledge distillation}, a novel form of knowledge distillation that uses the pairwise activation similarities within each input mini-batch  to supervise the training of a student network with a trained teacher network.
    \item We experimentally validate our approach on three public datasets. Our experiments show the potential of similarity-preserving knowledge distillation, not only for improving the training outcomes of student networks, but also for complementing traditional methods for knowledge distillation. 
\end{itemize}

\section{Method}
\label{sec:method}

\begin{figure*}
\centering
\parbox[b][2.0cm][t]{.17\linewidth}{WideResNet-16-1\\(0.2M params)}
\includegraphics[width=0.2\linewidth]{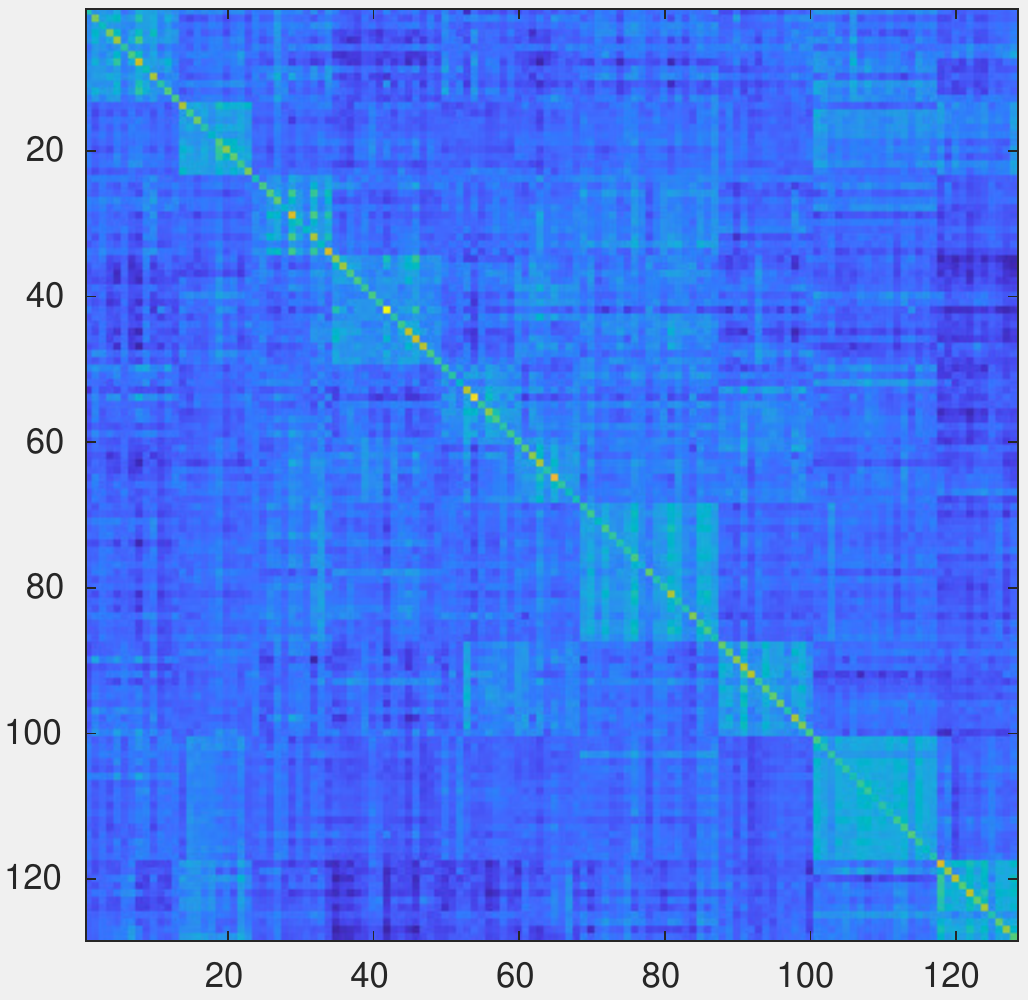}
\includegraphics[width=0.2\linewidth]{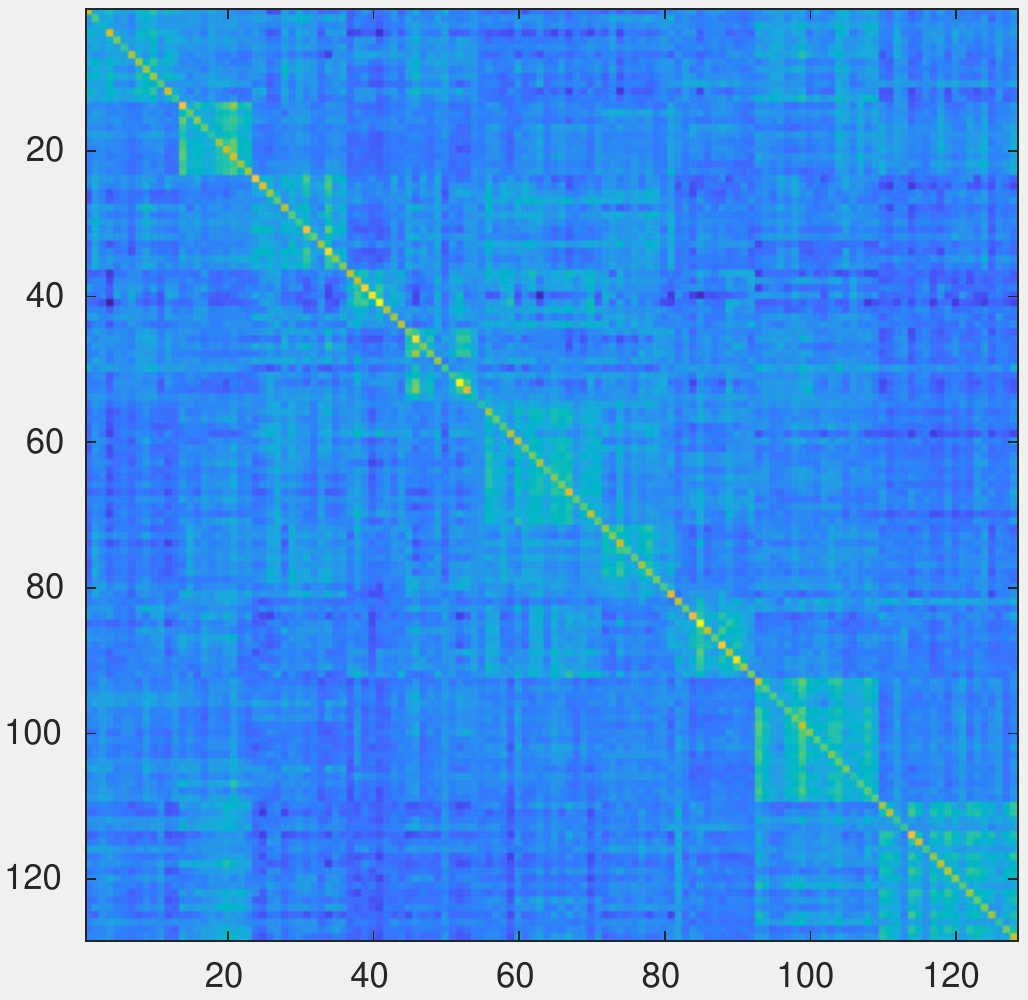}
\includegraphics[width=0.2\linewidth]{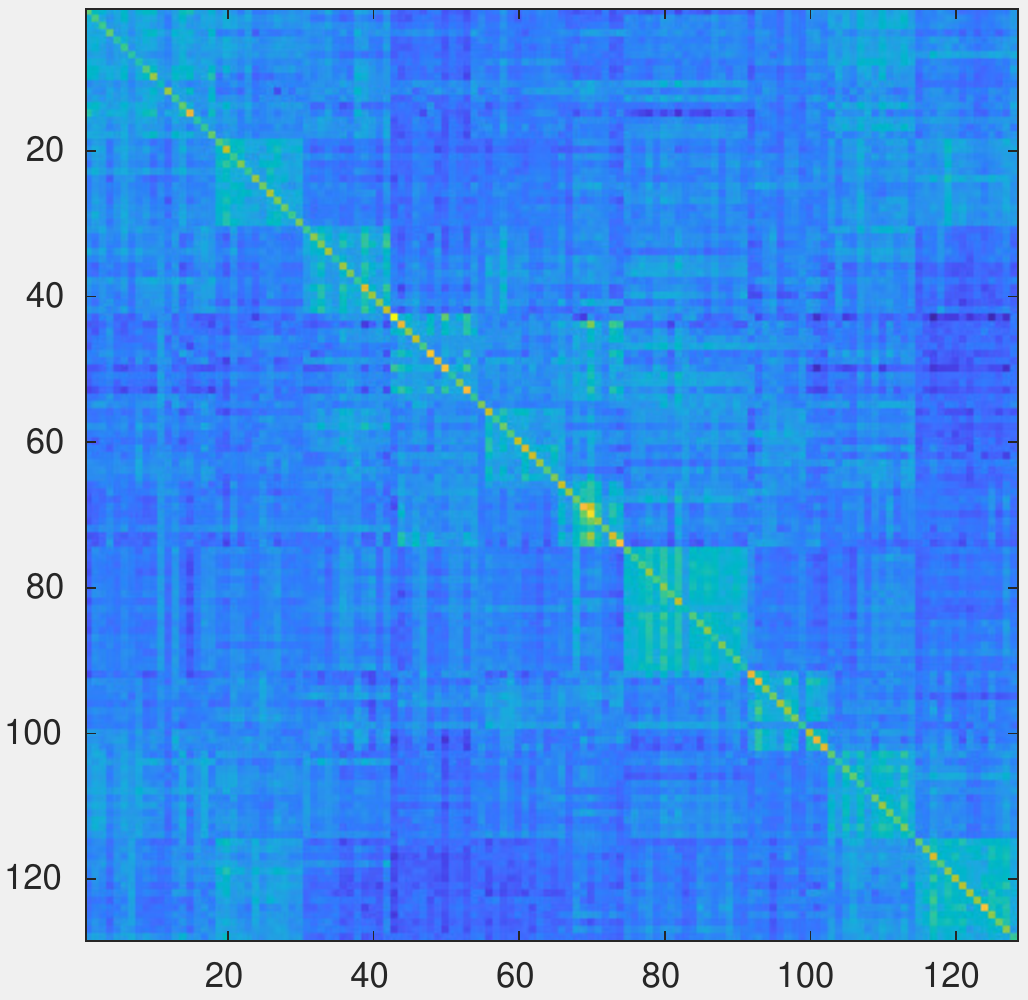}
\includegraphics[width=0.2\linewidth]{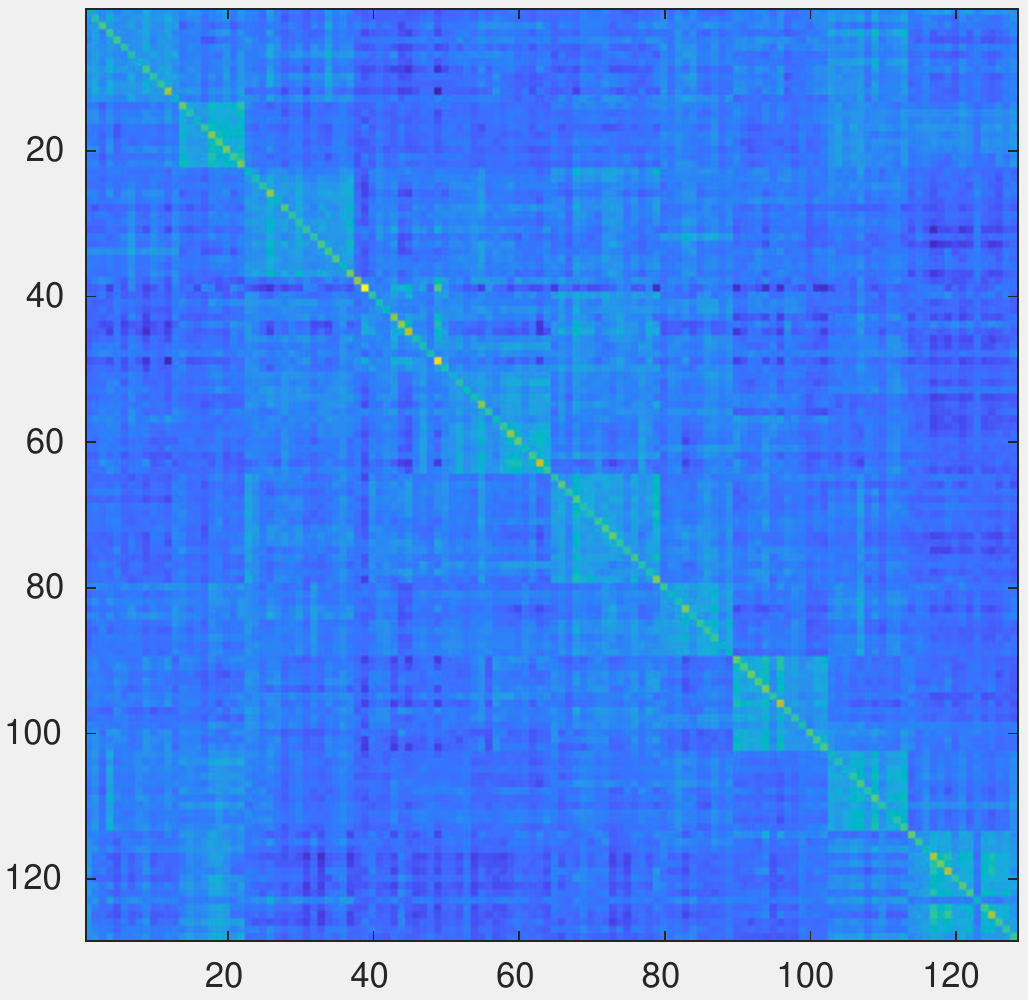}\\
\parbox[b][2.0cm][t]{.17\linewidth}{WideResNet-40-2\\(2.2M params)}
\includegraphics[width=0.2\linewidth]{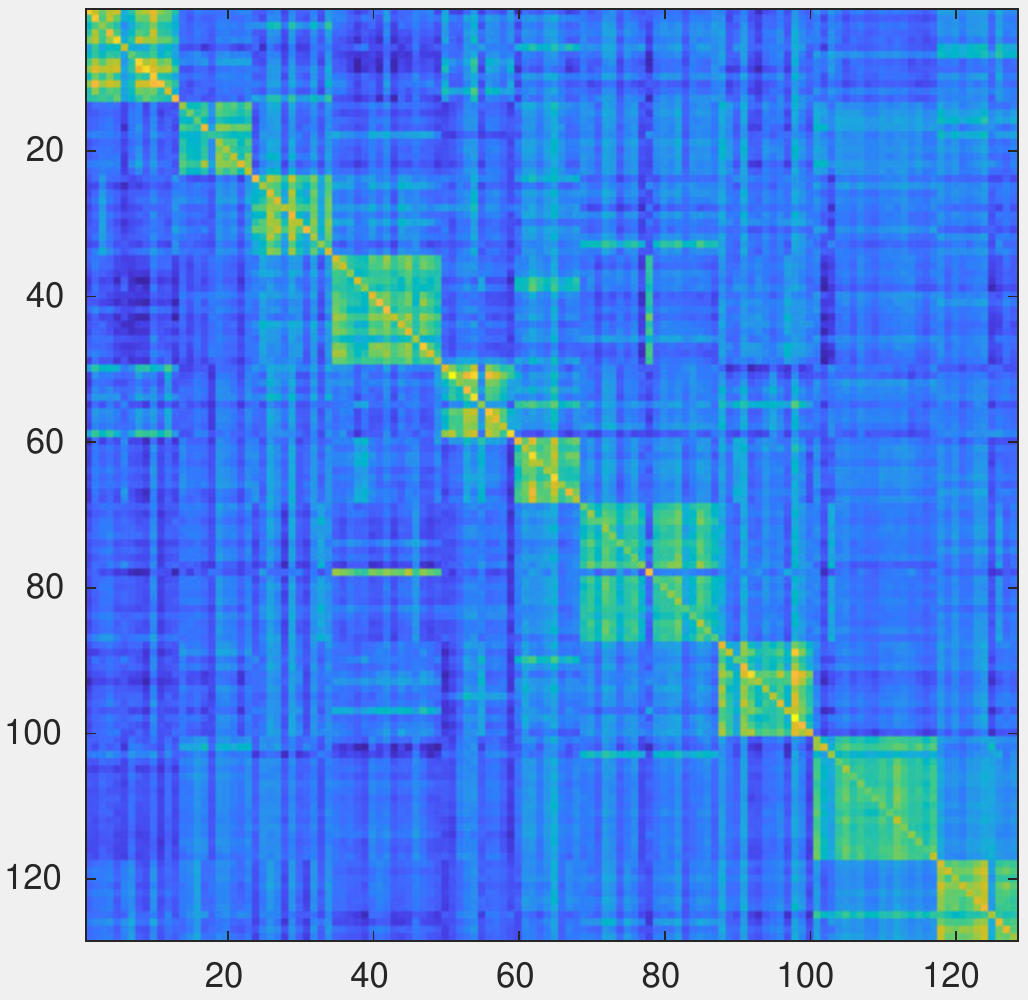}
\includegraphics[width=0.2\linewidth]{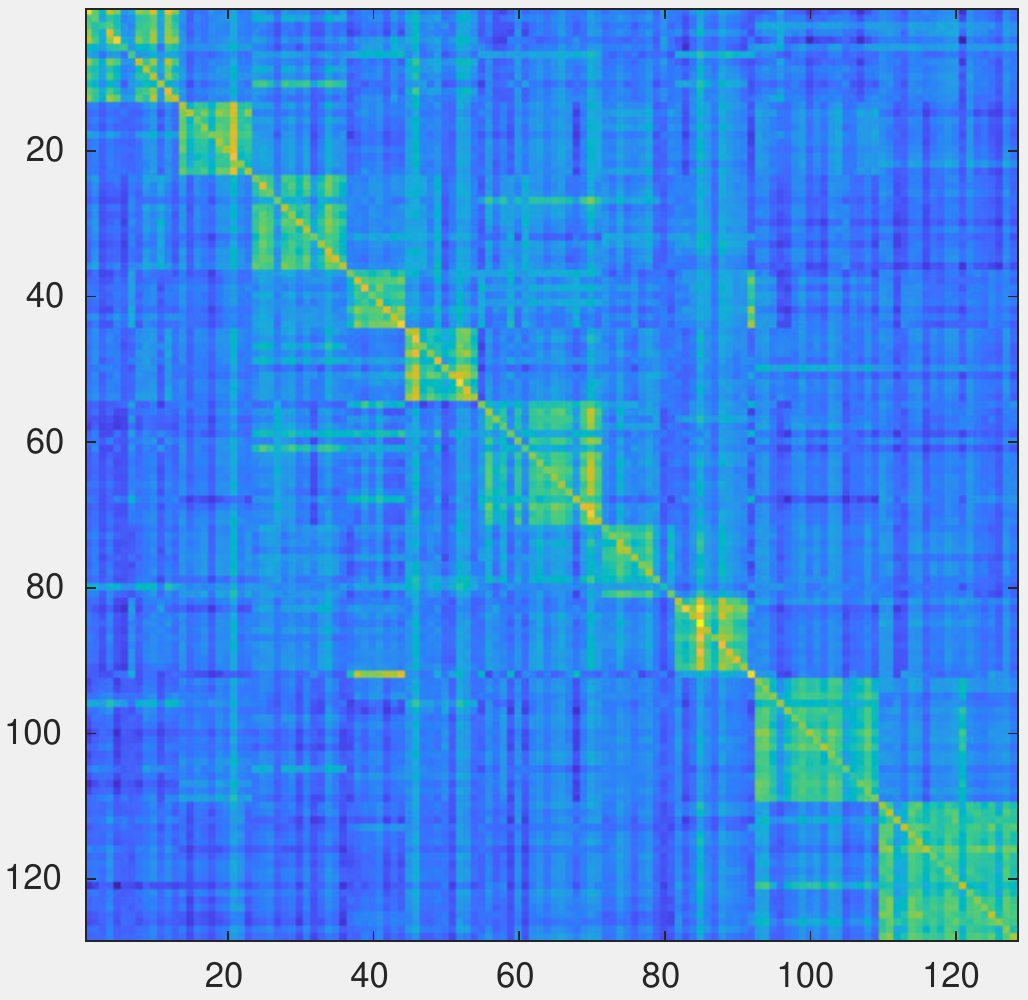}
\includegraphics[width=0.2\linewidth]{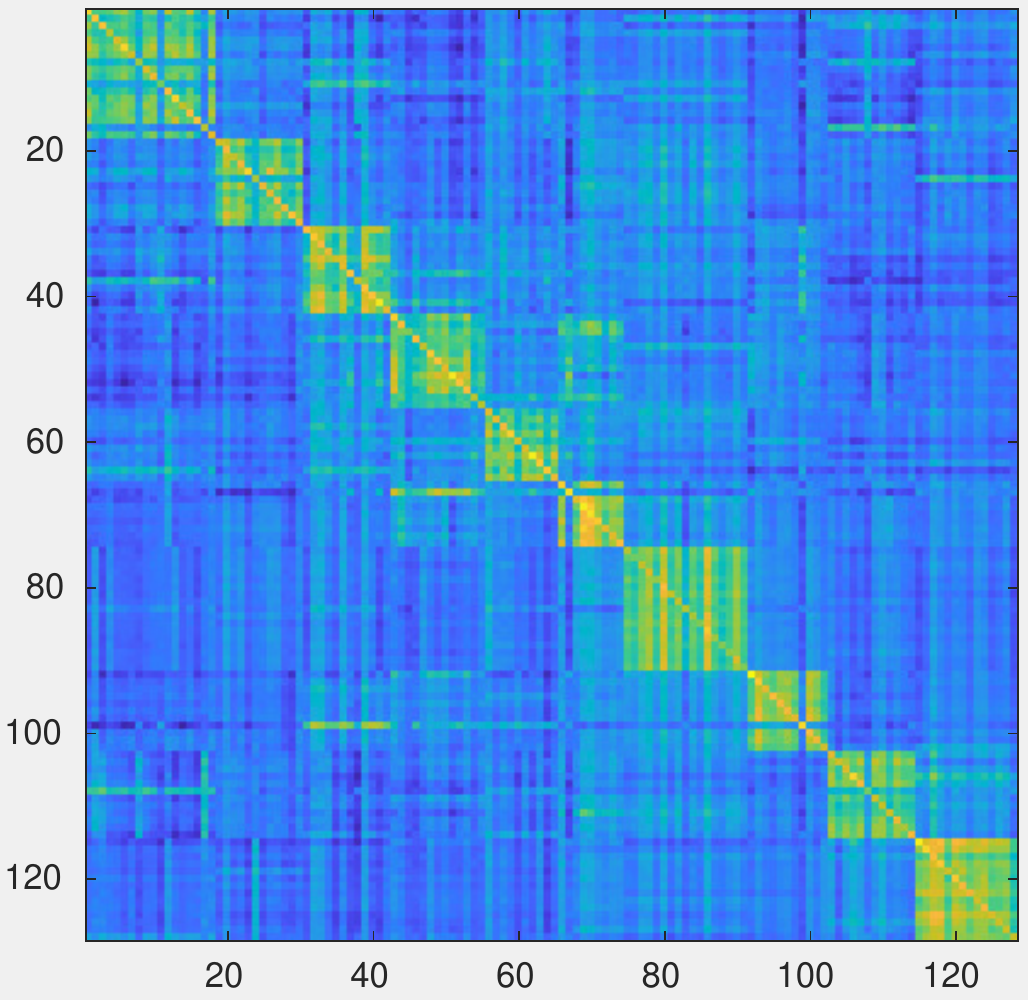}
\includegraphics[width=0.2\linewidth]{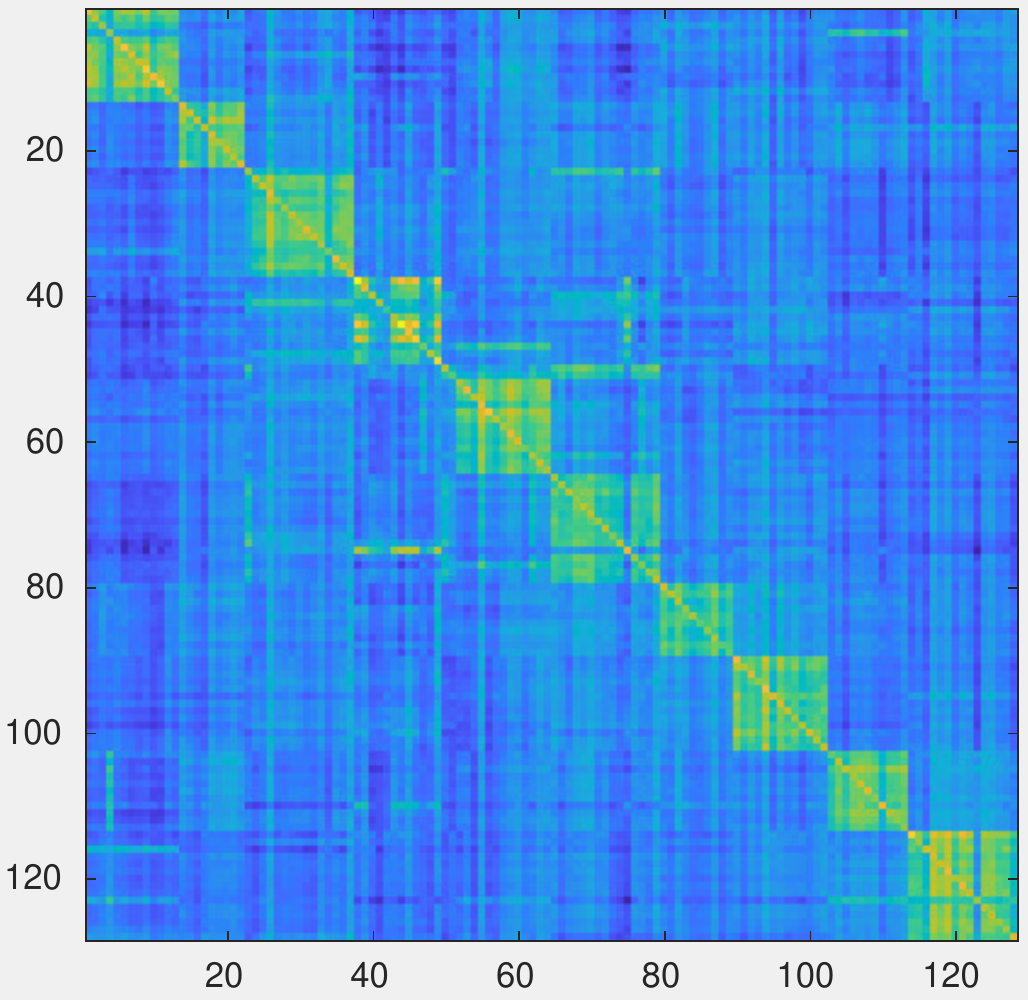}
\caption{Activation similarity matrices $G$ (Eq. \ref{eq:g}) produced by trained WideResNet-16-1 and WideResNet-40-2 networks on sample CIFAR-10 test batches. Each column shows a single batch with inputs grouped by ground truth class along each axis (batch size = 128). Brighter colors indicate higher similarity values. The blockwise patterns indicate that the elicited activations are mostly similar for inputs of the same class, and different for inputs across different classes. Our distillation loss (Eq. \ref{eq:bcloss}) encourages the student network to produce $G$ matrices closer to those produced by the teacher network.}
\label{fig:g}
\end{figure*}

The goal of knowledge distillation is to train a student network under the guidance of a trained teacher network, which acts as an extra source of supervision. For example, in neural network compression, the student network is computationally cheaper than the teacher: it may be shallower, thinner, or composed of cheaper operations. The trained teacher network provides additional semantic knowledge beyond the usual data supervision (e.g. the usual one-hot vectors for classification). The challenge is to determine how to encode and transfer the teacher's knowledge such that student performance is maximized.

In traditional knowledge distillation \cite{hintonetal2015}, knowledge is encoded and transferred in the form of softened class scores. The total loss for training the student is given by
\begin{equation}
\label{eq:kd}
\mathcal{L} = (1-\alpha) \mathcal{L}_{\text{CE}}(\mathbf{y}, \sigma(\mathbf{z}_S)) + 2 \alpha T^2 \mathcal{L}_{\text{CE}}(\sigma(\frac{\mathbf{z}_S}{T}), \sigma(\frac{\mathbf{z}_T}{T}))  \; ,
\end{equation}

\noindent where $\mathcal{L}_{\text{CE}}(\cdot,\cdot)$ denotes the cross-entropy loss, $\sigma(\cdot)$ denotes the softmax function, $\mathbf{y}$ is the one-hot vector indicating the ground truth class, $\mathbf{z}_S$ and $\mathbf{z}_T$ are the output logits of the student and teacher networks, respectively, $T$ is a temperature hyperparameter, and $\alpha$ is a balancing hyperparameter. The first term in Eq.\ref{eq:kd} is the usual cross-entropy loss defined using data supervision (ground truth labels), while the second term encourages the student to mimic the softened class scores of the teacher.

Recall from the introduction and Figure \ref{fig:activations} that semantically similar inputs tend to elicit similar activation patterns in a trained neural network. In Figure \ref{fig:activations}, we can observe that activation patterns are largely consistent within the same object category and distinctive across different categories. Might the correlations in activations encode useful teacher knowledge that can be transferred to the student? Our hypothesis is that, if two inputs produce highly similar activations in the teacher network, it is beneficial to guide the student network towards a configuration that also results in the two inputs producing highly similar activations in the student. Conversely, if two inputs produce dissimilar activations in the teacher, we want these inputs to produce dissimilar activations in the student as well.

Given an input mini-batch, denote the activation map produced by the teacher network $T$ at a particular layer $l$ by $A_T^{(l)} \in \mathbf{R}^{b \times c \times h \times w}$, where $b$ is the batch size, $c$ is the number of output channels, and $h$ and $w$ are spatial dimensions. Let the activation map produced by the student network $S$ at a corresponding layer $l'$ be given by $A_S^{(l')} \in \mathbf{R}^{b \times c' \times h' \times w'}$. Note that $c$ does not necessarily have to equal $c'$, and likewise for the spatial dimensions. Similar to attention transfer \cite{zagoruykokomodakis2017}, the corresponding layer $l'$ can be the layer at the same depth as $l$ if the student and teacher share the same depth, or the layer at the end of the same block if the student and teacher have different depths. To guide the student towards the activation correlations induced in the teacher, we define a distillation loss that penalizes differences in the L2-normalized outer products of $A_T^{(l)}$ and $A_S^{(l')}$. First, let 
\begin{equation}
\label{eq:g}
\rev{\tilde{G}_T^{(l)} = Q_T^{(l)} \cdot Q_T^{(l)\top} ; \;\;
G_{T[i,:]}^{(l)} = \tilde{G}_{T[i,:]}^{(l)} \, / \, ||\tilde{G}_{T[i,:]}^{(l)}||_2}
\end{equation}
\noindent where $Q_T^{(l)} \in \mathbf{R}^{b \times chw}$ is a reshaping of $A_T^{(l)}$, and therefore \rev{$\tilde{G}_T^{(l)}$} is a $b \times b$ matrix. Intuitively, entry $(i,j)$ in \rev{$\tilde{G}^{(l)}_T$} encodes the similarity of the activations at this teacher layer elicited by the $i$th and $j$th images in the mini-batch. \rev{We apply a row-wise L2 normalization to obtain $G_T^{(l)}$, where the notation $[i,:]$ denotes the $i$th row in a matrix.} Analogously, let
\begin{equation}
\rev{\tilde{G}_S^{(l)} = Q_S^{(l)} \cdot Q_S^{(l)\top} ; \;\;
G_{S[i,:]}^{(l)} = \tilde{G}_{S[i,:]}^{(l)} \, / \, ||\tilde{G}_{S[i,:]}^{(l)}||_2}
\end{equation}
\noindent where $Q_S^{(l')} \in \mathbf{R}^{b \times c'h'w'}$ is a reshaping of $A_S^{(l')}$, and $G_S^{(l')}$ is a $b \times b$ matrix. 
We define the similarity-preserving knowledge distillation loss as:
\begin{equation}
\label{eq:bcloss}
    \mathcal{L}_{\text{SP}}(G_T, G_S) = \frac{1}{b^2} \sum_{(l,l') \in \mathcal{I}}  ||G_T^{(l)} - G_S^{(l')}||^2_F \; ,
\end{equation}
where $\mathcal{I}$ collects the $(l,l')$ layer pairs (e.g. layers at the end of the same block, as discussed above) and $||\cdot||_F$ is the Frobenius norm. Eq. \ref{eq:bcloss} is a summation, over all $(l,l')$ pairs, of the mean element-wise squared difference between the $G_T^{(l)}$ and $G_S^{(l')}$ matrices. Finally, we define the total loss for training the student network as:
\begin{equation}
\label{eq:l}
    \mathcal{L} = \mathcal{L}_{\text{CE}}(\mathbf{y}, \sigma(\mathbf{z}_S)) + \gamma \, \mathcal{L}_{\text{SP}}(G_T, G_S) \; ,
\end{equation}
where $\gamma$ is a balancing hyperparameter.

\begin{table}
\centering
\small
\begin{tabular}{c|c|c|c}
Group & Output & WideResNet-16-$k$ & WideResNet-40-$k$\\
name & size & & \\
\hline
conv1 & $32 \times 32$ & $3 \times 3, 16$ &  $3 \times 3, 16$ \\
\hline
conv2 & $32 \times 32$ & $\begin{bmatrix} 3 \times 3, 16k \\ 3 \times 3, 16k \end{bmatrix} \times 2$ & $\begin{bmatrix} 3 \times 3, 16k \\ 3\times3, 16k \end{bmatrix} \times 6$ \\
\hline 
conv3 & $16 \times 16$ & $\begin{bmatrix} 3 \times 3, 32k \\ 3 \times 3, 32k \end{bmatrix} \times 2$ & $\begin{bmatrix} 3 \times 3, 32k \\ 3\times3, 32k \end{bmatrix} \times 6$ \\
\hline
conv4 & $8 \times 8$ & $\begin{bmatrix} 3 \times 3, 64k \\ 3 \times 3, 64k \end{bmatrix} \times 2$ & $\begin{bmatrix} 3 \times 3, 64k \\ 3\times3, 64k \end{bmatrix} \times 6$ \\
\hline
& $1 \times 1$ & \multicolumn{2}{c}{average pool, 10-d fc, softmax}\\
\hline
\noalign{\bigskip}
\end{tabular}
\caption{Structure of WideResNet networks used in CIFAR-10 experiments. Downsampling is performed by strided convolutions in the first layers of conv3 and conv4.}
\label{tab:cifar10nets}
\end{table}

\begin{table*}
\centering
\begin{tabular}{ll|cccc|c}
Student & Teacher & Student & KD \cite{hintonetal2015} & AT \cite{zagoruykokomodakis2017} & SP (ours) & Teacher\\
\hline
WideResNet-16-1 (0.2M) & WideResNet-40-1 (0.6M) & 8.74 & 8.48 & 8.30 & \textbf{8.13} & 6.51\\
WideResNet-16-1 (0.2M) & WideResNet-16-2 (0.7M) & 8.74 & 7.94 & 8.28 & \textbf{7.52} & 6.07\\
WideResNet-16-2 (0.7M) & WideResNet-40-2 (2.2M) & 6.07 & 6.00 & 5.89 & \textbf{5.52} & 5.18 \\
WideResNet-16-2 (0.7M) & WideResNet-16-8 (11.0M) & 6.07 & 5.62 & 5.47 & \textbf{5.34} & 4.24 \\
WideResNet-40-2 (2.2M) & WideResNet-16-8 (11.0M) & 5.18 & 4.86 & \textbf{4.47} & 4.55 & 4.24 \\
\hline 
\smallskip
\end{tabular}
\caption{Experiments on CIFAR-10 with three different knowledge distillation losses: softened class scores (traditional KD), attention transfer (AT), and similarity preserving (SP). The median error over five runs is reported, following the protocol in \cite{zagoruykokomodakis2016,zagoruykokomodakis2017}. The best result for each experiment is shown in bold. Brackets indicate model size in number of parameters.}
\label{tab:cifar10}
\end{table*}

Figure \ref{fig:g} visualizes the $G$ matrices for several batches in the CIFAR-10 test set. The top row is produced by a trained WideResNet-16-1 network, consisting of 0.2M parameters, while the bottom row is produced by a trained WideResNet-40-2 network, consisting of 2.2M parameters. In both cases, activations are collected from the last convolution layer. Each column represents a single batch, which is identical for both networks. The images in each batch have been grouped by their ground truth class for easier interpretability. The $G$ matrices in both rows show a distinctive blockwise pattern, indicating that the activations at the last layer of these networks are largely similar within the same class and dissimilar across different classes (the blocks are differently sized because each batch has an unequal number of test samples from each class). Moreover, the blockwise pattern is more distinctive for the WideResNet-40-2 network, reflecting the higher capacity of this network to capture the semantics of the dataset. Intuitively, Eq. \ref{eq:bcloss} pushes the student network towards producing $G$ matrices closer to those produced by the teacher network.

\textbf{Differences from previous approaches.} The similarity-preserving knowledge distillation loss (Eq. \ref{eq:bcloss}) is defined in terms of activations instead of class scores as in traditional distillation \cite{hintonetal2015}. Activations are also used to define the distillation losses in FitNets \cite{romeroetal2015}, flow-based distillation \cite{yimetal2017}, and attention transfer \cite{zagoruykokomodakis2017}. However, a key difference is that these
previous distillation methods encourage the student to mimic different aspects of the representation space of the teacher. Our method is a departure from this common approach in that it aims to preserve the pairwise activation similarities of input samples. Its behavior is unchanged by a rotation of the teacher's representation space, for example. In similarity-preserving knowledge distillation, the student is not required to be able to express the representation space of the teacher, as long as pairwise similarities in the teacher space are well preserved in the student space.

\section{Experiments}

We now turn to the experimental validation of our distillation approach on three public datasets. We start with CIFAR-10 as it is a commonly adopted dataset for comparing distillation methods, and its relatively small size allows multiple student and teacher combinations to be evaluated. We then consider the task of transfer learning, and show how distillation and fine-tuning can be combined to perform transfer learning on a texture dataset with limited training data. Finally, we report results on the larger CINIC-10 dataset.

\subsection{CIFAR-10}

CIFAR-10 consists of 50,000 training images and 10,000 testing images at a resolution of 32x32. The dataset covers ten object classes, with each class having an equal number of images.
We conducted experiments using wide residual networks (WideResNets) \cite{zagoruykokomodakis2016} following 
\cite{crowleyetal2018,zagoruykokomodakis2017}. 
Table \ref{tab:cifar10nets} summarizes the structure of the networks.
We adopted the standard protocol \cite{zagoruykokomodakis2016} for training wide residual networks on CIFAR-10 (SGD with Nesterov momentum; 200 epochs; batch size of 128; and an initial learning rate of 0.1, decayed by a factor of 0.2 at epochs 60, 120, and 160). We applied the standard horizontal flip and random crop data augmentation.
We performed baseline comparisons with respect to traditional knowledge distillation (softened class scores) and attention transfer.
For traditional knowledge distillation \cite{hintonetal2015}, we set $\alpha=0.9$ and $T=4$ following the CIFAR-10 experiments in \cite{crowleyetal2018,zagoruykokomodakis2017}. Attention transfer losses were applied for each of the three residual block groups. We set the weight of the distillation loss in attention transfer and similarity-preserving distillation by held-out validation on a subset of the training set ($\beta=1000$ for attention transfer, $\gamma=3000$ for similarity-preserving distillation).

\begin{figure*}
\centering
\includegraphics[width=0.32\linewidth]{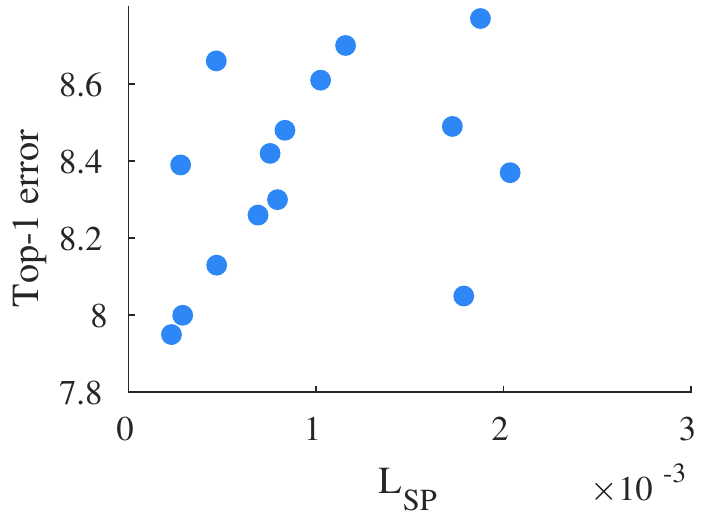}
\includegraphics[width=0.32\linewidth]{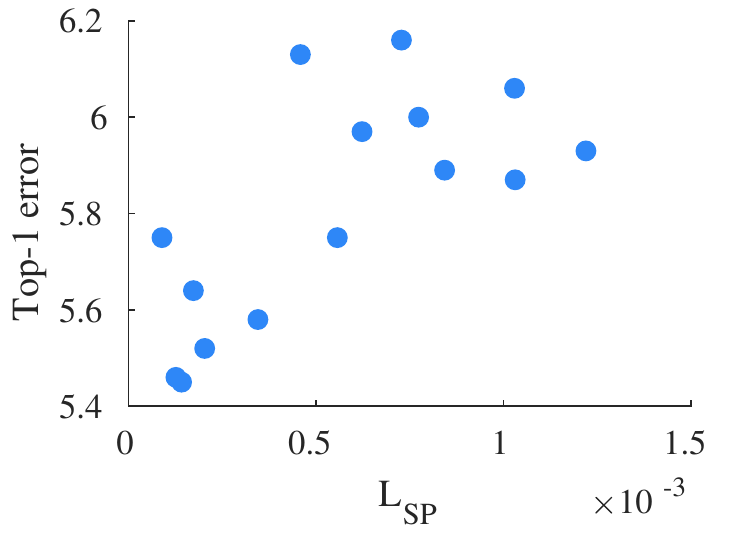}
\includegraphics[width=0.32\linewidth]{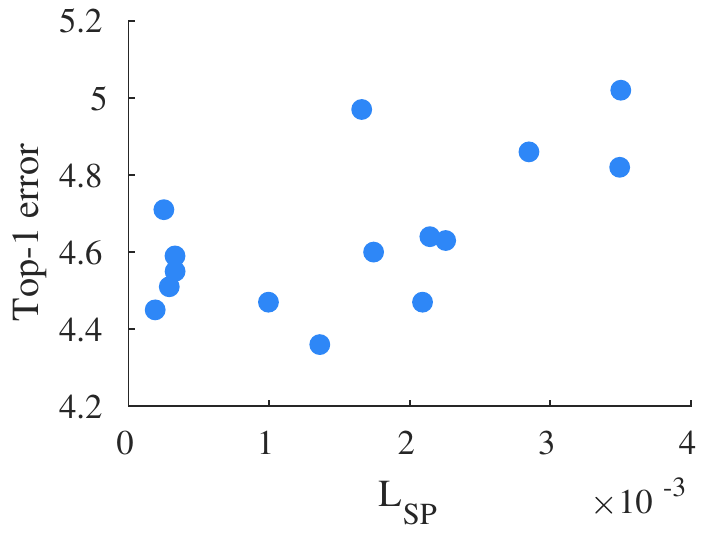}
\caption{$\mathcal{L}_{\text{SP}}$ vs. error for (from left to right) WideResNet-16-1 students trained with WideResNet-16-2 teachers, WideResNet-16-2 students trained with WideResNet-40-2 teachers, and WideResNet-40-2 students trained with WideResNet-16-8 teachers, on CIFAR-10.}
\label{fig:la}
\end{figure*}

\begin{table*}
\centering
\begin{tabular}{ll|ccc|c}
Student & Teacher & Student & AT \cite{zagoruykokomodakis2017} & SP  (win:loss) & Teacher\\
\hline
MobileNet-0.25 (0.2M) & MobileNet-0.5 (0.8M) & 42.45 & 42.39 & \textbf{41.30} \, (7:3) & 36.76\\
MobileNet-0.25 (0.2M) & MobileNet-1.0 (3.3M) & 42.45 & 41.89 & \textbf{41.76} \, (5:5) & 34.10\\
MobileNet-0.5 (0.8M) & MobileNet-1.0 (3.3M) & 36.76 & 35.61 & \textbf{35.45} \, (7:3) & 34.10\\
\hline
MobileNetV2-0.35 (0.5M) & MobileNetV2-1.0 (2.2M) & 41.25 & 41.60 & \textbf{40.29} \, (8:2) & 36.62 \\
MobileNetV2-0.35 (0.5M) & MobileNetV2-1.4 (4.4M) & 41.25 & 41.04 & \textbf{40.43} \, (8:2) & 35.35 \\
MobileNetV2-1.0 (2.2M) & MobileNetV2-1.4 (4.4M) & 36.62 & 36.33 & \textbf{35.61} \, (8:2) & 35.35\\
\hline
\smallskip
\end{tabular}
\caption{Transfer learning experiments on the describable textures dataset with attention transfer (AT) and similarity preserving (SP) knowledge distillation. The median error over the ten standard splits is reported. The best result for each experiment is shown in bold. The (win:loss) notation indicates the number of splits in which SP outperformed AT. The (*M) notation indicates model size in number of parameters.}
\label{tab:dtd}
\end{table*}

Table \ref{tab:cifar10} shows our results experimenting with several student-teacher network pairs. We tested cases in which the student and teacher networks have the same width but different depth (WideResNet-16-1 student with WideResNet-40-1 teacher; WideResNet-16-2 student with WideResNet-40-2 teacher), the student and teacher networks have the same depth but different width (WideResNet-16-1 student with WideResNet-16-2 teacher; WideResNet-16-2 student with WideResNet-16-8 teacher), and the student and teacher have different depth and width (WideResNet-40-2 student with WideResNet-16-8 teacher). In all cases, transferring the knowledge of the teacher network using similarity-preserving distillation improved student training outcomes. Compared to conventional training with data supervision (i.e. one-hot vectors), the student network consistently obtained lower median error, from 0.5 to 1.2 absolute percentage points, or 7\% to 14\% relative, with no additional network parameters or operations.
Similarity-preserving distillation also performed favorably with respect to the traditional (softened class scores) and attention transfer baselines, achieving the lowest error in four of the five cases. This validates our intuition that the activation similarities across images encode useful semantics learned by the teacher network, and provide an effective supervisory signal for knowledge distillation.

Figure \ref{fig:la} plots $\mathcal{L}_{\text{SP}}$ vs. error for the WideResNet-16-1/WideResNet-16-2, WideResNet-16-2/WideResNet-40-2, and WideResNet-40-2/WideResNet-16-8 experiments (left to right, respectively), using all students trained with traditional KD, AT, and SP. The plots verify that $\mathcal{L}_{\text{SP}}$ and performance are correlated.

While we have presented these results from the perspective of improving the training of a student network, it is also possible to view the results from the perspective of the teacher network. Our results suggest the potential for using similarity-preserving distillation to compress large networks into more resource-efficient ones with minimal accuracy loss. In the fifth test, for example, the knowledge of a trained WideResNet-16-8 network, which contains 11.0M parameters, is distilled into a much smaller WideResNet-40-2 network, which contains only 2.2M parameters. This is a $5\times$ compression rate with only $0.3\%$ loss in accuracy, using off-the-shelf PyTorch without any specialized hardware or software.

The above similarity-preserving distillation results were produced using only the activations collected from the last convolution layers of the student and teacher networks. We also experimented with using the activations at the end of each WideResNet block, but found no improvement in performance. We therefore used only the activations at the final convolution layers in the subsequent experiments. Activation similarities may be less informative in the earlier layers of the network because these layers encode more generic features, which tend to be present across many images. Progressing deeper in the network, the channels encode increasingly specialized features, and the activation patterns of semantically similar images become more distinctive.
We also experimented with using post-softmax scores to determine similarity, but this produces worse results than using activations. We found the same when using an oracle, suggesting that the soft teacher signal is important.

\subsection{Transfer learning combining distillation with fine-tuning}

In this section, we explore a common transfer learning scenario in computer vision. Suppose we are faced with a novel recognition task in a specialized image domain with limited training data. A natural strategy to adopt is to transfer the knowledge of a network pre-trained on ImageNet (or another suitable large-scale dataset) to the new recognition task by fine-tuning. Here, we combine knowledge distillation with fine-tuning: we initialize the student network with source domain (in this case, ImageNet) pretrained weights, and then fine-tune the student to the target domain using both distillation and cross-entropy losses (Eq. \ref{eq:l}). 

We analyzed this scenario using the describable textures dataset \cite{cimpoietal2014}, which is composed of 5,640 images covering 47 texture categories. Image sizes range from 300x300 to 640x640. We applied ImageNet-style data augmentation with horizontal flipping and random resized cropping during training. At test time, images were resized to 256x256 and center cropped to 224x224 for input to the networks. For evaluation, we adopted the standard ten training-validation-testing splits. 
To demonstrate the versatility of our method on different network architectures, and in particular its compatibility with mobile-friendly architectures, we experimented with variants of MobileNet \cite{howardetal2017} and MobileNetV2 \cite{sandleretal2018}. Tables 1 and 2 in the supplementary summarize the structure of the networks.

We compared with an attention transfer baseline. Softened class score based distillation is not directly comparable in this setting because the classes in the source and target domains are disjoint. 
Similarity-preserving distillation can be applied directly to train the student, without first fine-tuning the teacher, since it aims to preserve similarities instead of mimicking the teacher's representation space. 
The teacher is run in inference mode to generate representations in the new domain. This capacity is useful when the new domain has limited training data, when the source domain is not accessible to the student (e.g. in privileged learning), or in continual learning where trained knowledge needs to be preserved across tasks~\footnote{In continual (or lifelong) learning, the goal is to extend a trained network to new tasks while avoiding the catastrophic forgetting of previous tasks. One way to prevent catastrophic forgetting is to supervise the new model (the student) with the model trained for previous tasks (the teacher) via a knowledge distillation loss \cite{zhaietal2019}.}.
We set the hyperparameters for attention transfer and similarity-preserving distillation by held-out validation on the ten standard splits. All networks were trained using SGD with Nesterov momentum, a batch size of 96, and for 60 epochs with an initial learning rate of 0.01 reduced to 0.001 after 30 epochs.

Table \ref{tab:dtd} shows that similarity-preserving distillation can effectively transfer knowledge across different domains. For all MobileNet and MobileNetV2 student-teacher pairs tested, applying similarity-preserving distillation during fine-tuning resulted in lower median student error than fine-tuning without distillation. Fine-tuning MobileNet-0.25 with distillation reduced the error by 1.1\% absolute, and fine-tuning MobileNet-0.5 with distillation reduced the error by 1.3\% absolute, compared to fine-tuning without distillation. Fine-tuning MobileNetV2-0.35 with distillation reduced the error by 1.0\% absolute, and fine-tuning MobileNetV2-1.0 with distillation reduced the error by 1.0\% absolute, compared to fine-tuning without distillation. 

For all student-teacher pairs, similarity-preserving distillation obtained lower median error than the spatial attention transfer baseline. Table \ref{tab:dtd} incudes a breakdown of how similarity-preserving distillation compares with spatial attention transfer on a per-split basis. On aggregate, similarity-preserving distillation outperformed spatial attention transfer on 19 out of the 30 MobileNet splits and 24 out of the 30 MobileNetV2 splits. The results suggest that there may be a challenging domain shift in the important image areas for the network to attend. Moreover, while attention transfer summarizes the activation map by summing out the channel dimension, similarity-preserving distillation makes use of the full activation map in computing the similarity-based distillation loss, which may be more robust in the presence of a domain shift in attention. 

\subsection{CINIC-10} 
The CINIC-10 dataset \cite{darlowetal2018} is designed to be a middle option relative to CIFAR-10 and ImageNet: it is composed of 32x32 images in the style of CIFAR-10, but at a total of 270,000 images its scale is closer to that of ImageNet. We adopted CINIC-10 for rapid experimentation because several GPU-months would have been required to perform full held-out validation and training on ImageNet for our method and all baselines.

\begin{table*}
\centering
\begin{tabular}{ll|cccccc|c}
Student & Teacher & Student & KD \cite{hintonetal2015} & AT \cite{zagoruykokomodakis2017} & SP (ours) & KD+SP & AT+SP & Teacher\\
\hline
Sh.NetV2-0.5 (0.4M) & Sh.NetV2-1.0 (1.3M) & 20.09 & 18.62 & 18.50 & 18.56 & 18.35 & \textbf{18.20} & 17.26 \\
Sh.NetV2-0.5 (0.4M) & Sh.NetV2-2.0 (5.3M) & 20.09 & 18.96 & 18.78 & 19.09 & 18.88 & \textbf{18.43} & 15.63 \\
Sh.NetV2-1.0 (1.3M) & Sh.NetV2-2.0 (5.3M) & 17.26 & 16.01 & 15.95 & 15.95 & 16.11 & \textbf{15.89} & 15.63 \\
\hline
M.NetV2-0.35 (0.4M) & M.NetV2-1.0 (2.2M) & 17.12 & 16.27 & 16.10 & 16.57 & 16.17 & \textbf{15.66} & 14.05 \\
\hline
\smallskip
\end{tabular}
\caption{Experiments on CINIC-10 with three different knowledge distillation losses: softened class scores (traditional KD), attention transfer (AT), and similarity preserving (SP). The best result for each experiment is shown in bold. Brackets indicate model size in number of parameters.}
\label{tab:cinic10}
\end{table*}

For the student and teacher architectures, we experimented with variants of the state-of-the-art mobile architecture ShuffleNetV2 \cite{maetal2018}. The ShuffleNetV2 networks are summarized in Table 
3 in the supplementary. We used the standard training-validation-testing split and set the hyperparameters for similarity-preserving distillation and all baselines by held-out validation (KD: $\{\alpha=0.6, T=16\}$; AT: $\beta=50$; SP: $\gamma=2000$; KD+SP: $\{\alpha=0.6,T=16,\gamma=2000\}$; AT+SP: $\{\beta=30,\gamma=2000\}$). All networks were trained using SGD with Nesterov momentum, a batch size of 96, for 140 epochs with an initial learning rate of 0.01 decayed by a factor of 10 after the 100th and 120th epochs. We applied CIFAR-style data augmentation with horizontal flips and random crops during training.

The results are shown in Table \ref{tab:cinic10} (top). Compared to conventional training with data supervision only, similarity-preserving distillation consistently improved student training outcomes. In particular, training ShuffleNetV2-0.5 with similarity-preserving distillation reduced the error by 1.5\% absolute, and training ShuffleNetV2-1.0 with similarity-preserving distillation reduced the error by 1.3\% absolute.
On an individual basis, all three knowledge distillation approaches achieved comparable results, with a total spread of 0.12\% absolute error on ShuffleNetV2-0.5 (for the best results with ShuffleNetV2-1.0 as teacher) and a total spread of 0.06\% absolute error on ShuffleNetV2-1.0. However, the lowest error was achieved by combining similarity-preserving distillation with spatial attention transfer. Training ShuffleNetV2-0.5 combining both distillation losses reduced the error by 1.9\% absolute, and training ShuffleNetV2-1.0 combining both distillation losses reduced the error by 1.4\% absolute. This result shows that similarity-preserving distillation complements attention transfer and captures teacher knowledge that is not fully encoded in spatial attention maps.
Table \ref{tab:cinic10} (bottom) summarizes additional experiments with MobileNetV2. The results are similar: SP does not outperform the individual baselines but complements traditional KD and AT.

\begin{figure}
\centering
\includegraphics[width=\linewidth]{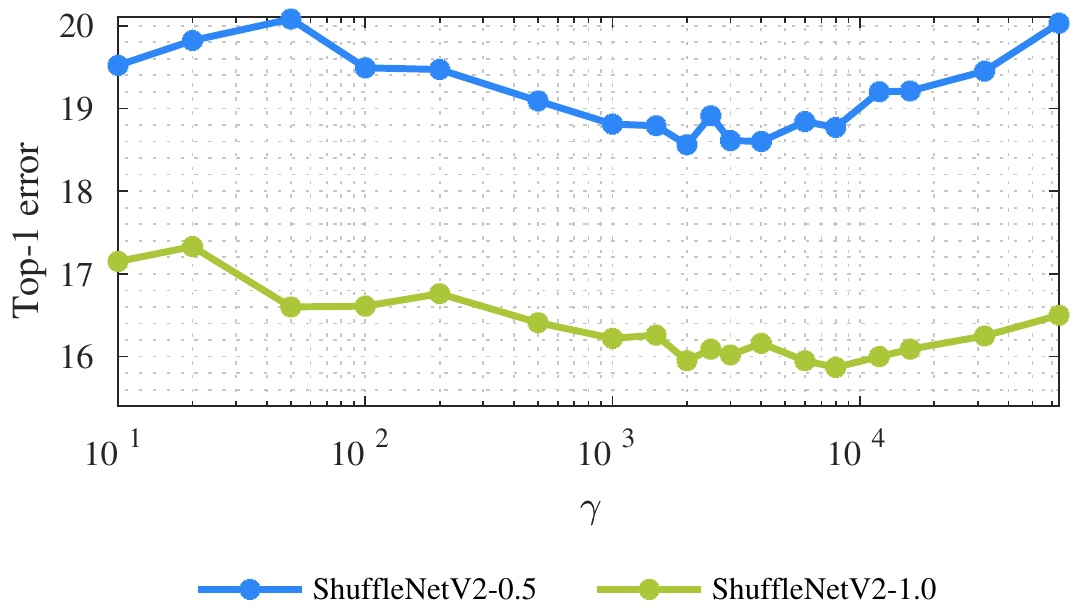}
\smallskip
\caption{Sensitivity to $\gamma$ on the CINIC-10 test set for ShuffleNetV2 students.}
\label{fig:varygamma}
\end{figure}

\textbf{Sensitivity analysis.} Figure \ref{fig:varygamma} illustrates how the performance of similarity-preserving distillation is affected by the choice of hyperparameter $\gamma$. We plot the top-1 errors on the CINIC-10 test set for ShuffleNetV2-0.5 and ShuffleNetV2-1.0 students trained with $\gamma$ ranging from 
10 to 64,000. We observed robust performance over a broad range of values for $\gamma$. In all experiments, we set $\gamma$ by held-out validation.

\subsection{Different student and teacher architectures}

\begin{table*}
\centering
\begin{tabular}{ll|cccc|c}
Student & Teacher & Student & KD \cite{hintonetal2015} & AT \cite{zagoruykokomodakis2017} & SP (ours) & Teacher\\
\hline
ShuffleNetV2-0.5 (0.4M) & WideResNet-40-2 (2.2M) & 8.95 & 9.09 & 8.55 & \textbf{8.19} & 5.18 \\
MobileNetV2-0.35 (0.4M) & WideResNet-40-2 (2.2M) & 7.44 & 7.08 & 7.43 & \textbf{6.62} & 5.18 \\
\hline 
\smallskip
\end{tabular}
\caption{Additional experiments  with students and teachers from different architecture families on CIFAR-10. The median error over five runs is reported, following the protocol in \cite{zagoruykokomodakis2016,zagoruykokomodakis2017}. The best result for each experiment is shown in bold. Brackets indicate model size in number of parameters.}
\label{tab:diffarch}
\end{table*}

We performed additional experiments with students and teachers from different architecture families on CIFAR-10. Table \ref{tab:diffarch} shows that, for both MobileNetV2 and ShuffleNetV2, SP outperforms conventional training as well as the traditional KD and AT baselines.

\section{Related Work}
\label{sec:related}

We presented in this paper a novel distillation loss for capturing and transferring knowledge from a teacher network to a student network. Several prior alternatives \cite{hintonetal2015,romeroetal2015,yimetal2017,zagoruykokomodakis2017} are described in the introduction and some key differences are highlighted in Section \ref{sec:method}.
In addition to the knowledge capture (or loss definition) aspect of distillation studied in this paper, another important open question is the architectural design of students and teachers. In most studies of knowledge distillation, including ours, the student network is a thinner and/or shallower version of the teacher network. Inspired by efficient architectures such as MobileNet and ShuffleNet, Crowley et al. \cite{crowleyetal2018} proposed to replace regular convolutions in the teacher network with cheaper grouped and pointwise convolutions in the student. Ashok et al. \cite{ashoketal2018} developed a reinforcement learning approach to learn the student architecture. Polino et al. \cite{polinoetal2018} demonstrated how a quantized student network can be trained using a full-precision teacher network. 

There is also innovative orthogonal work exploring alternatives to the usual student-teacher training paradigm. Wang et al. \cite{wangetal2018} introduced an additional discriminator network, and trained the student, teacher, and discriminator networks together using a combination of distillation and adversarial losses. Lan et al. \cite{lanetal2018} proposed the on-the-fly native ensemble teacher model, in which the teacher is trained together with multiple students in a multi-branch network architecture. The teacher prediction is a weighted average of the branch predictions. 

Knowledge distillation was first introduced as a technique for neural network compression. Resource efficiency considerations have led to a recent increase in interest in efficient neural architectures \cite{howardetal2017,iandolaetal2016,maetal2018,sandleretal2018,zhangetal2018shufflenet}, as well as in algorithms for compressing trained deep networks.
Weight pruning methods \cite{hanetal2016,liuetal2018,luoetal2017,narangetal2017,tungmori2018,wenetal2016,yuetal2018} remove unimportant weights from the network, sparsifying the network connectivity structure. The induced sparsity is unstructured when individual connections are pruned, or structured when entire channels or filters are pruned. Unstructured sparsity usually results in better accuracy but requires specialized sparse matrix multiplication libraries \cite{skimcaffe} or hardware engines \cite{hanetal2016b} in practice.
Quantized networks \cite{faraoneetal2018,jacobetal2018,khoramli2018,rastegarietal2016,zhangetal2018,zhouetal2018}, such as fixed-point, binary, ternary, and arbitrary-bit networks, encode weights and/or activations using a small number of bits, or at lower precision.
Fractional or arbitrary-bit quantization \cite{frommetal2018,khoramli2018} encodes individual weights at different precisions, allowing multiple precisions to be used within a single network layer. 
Low-rank factorization methods \cite{dentonetal2014,dubeyetal2018,jaderbergetal2014,pengetal2018,zhangetal2015} produce compact low-rank approximations of filter matrices. 
Techniques from different categories have also been optimized jointly or combined sequentially to achieve higher compression rates \cite{dubeyetal2018,hanetal2016,tungmori2018}.

State-of-the-art network compression methods can achieve significant reductions in network size, in some cases by an order of magnitude, but often require specialized software or hardware support. For example, unstructured pruning requires optimized sparse matrix multiplication routines to realize practical acceleration \cite{skimcaffe}, platform constraint-aware compression \cite{chenetal2018,yangetal2017,yangetal2018} requires hardware simulators or empirical measurements, and arbitrary-bit quantization \cite{frommetal2018,khoramli2018} requires specialized hardware. One of the advantages of knowledge distillation is that it is easily implemented in any off-the-shelf deep learning framework without the need for extra software or hardware. Moreover, distillation can be integrated with other network compression techniques for further gains in performance \cite{polinoetal2018}.

\section{Conclusion}

We proposed similarity-preserving knowledge distillation: a novel form of knowledge distillation that aims to preserve pairwise similarities in the student's representation space, instead of mimicking the teacher's representation space. Our experiments demonstrate the potential of similarity-preserving distillation in improving the training outcomes of student networks compared to training with only data supervision (e.g. ground truth labels). Moreover, in a transfer learning setting, when traditional class score based distillation is not directly applicable, we have shown that similarity-preserving distillation provides a robust solution to the challenging domain shift problem. We have also shown that similarity-preserving distillation complements the state-of-the-art attention transfer method and captures teacher knowledge that is not fully encoded in spatial attention maps.
We believe that similarity-preserving distillation can provide a simple yet effective drop-in replacement for (or complement to) traditional forms of distillation in a variety of application areas, including model compression \cite{polinoetal2018}, privileged learning \cite{lopezpazetal2016}, adversarial defense \cite{papernotetal2016}, and learning with noisy data \cite{lietal2017}.

\textbf{Future directions.} As future work, we plan to explore similarity-preserving knowledge distillation in semi-supervised and omni-supervised \cite{radosavovicetal2018} learning settings. Since similarity-preserving distillation does not require labels, it is possible to distill further knowledge from the teacher using auxiliary images without annotations. For example, the supervised loss (e.g. cross-entropy) can be computed using the usual annotated training set, while the distillation loss can be computed using an auxiliary set of unlabelled web images. In this setting, the distillation loss is analogous to the reconstruction or unsupervised loss in semi-supervised learning.

{\small
\bibliographystyle{ieee}
\bibliography{egbib}
}

\end{document}